\def\year{2019}\relax
\documentclass[letterpaper]{article} %DO NOT CHANGE THIS
\usepackage{aaai19}  %Required
\usepackage{times}  %Required
\usepackage{helvet}  %Required
\usepackage{courier}  %Required
\usepackage{url}  %Required
\usepackage{graphicx}  %Required
\usepackage{multirow}
\usepackage{array}
\usepackage{amsfonts}
\usepackage{amsmath,bm}

\begin{document}
\title{A Two-Stream Mutual Attention Network for Semi-supervised Biomedical Segmentation with Noisy Labels}
\author{Shaobo Min, Xuejin Chen\thanks{Corresponding Author.}, Zheng-Jun Zha, Feng Wu, Yongdong Zhang\\
National Engineering Laboratory for Brain-inspired Intelligence Technology and Application,\\
University of Science and Technology of China, Hefei, Anhui, China\\ %  Removed for anonymous submission
}
\maketitle
\begin{abstract}
Learning-based methods suffer from a deficiency of clean annotations, especially in biomedical segmentation.
Although many semi-supervised methods have been proposed to provide extra training data, automatically generated labels are usually too noisy to retrain models effectively.
In this paper, we propose a Two-Stream Mutual Attention Network (TSMAN) that weakens the influence of back-propagated gradients caused by incorrect labels, thereby rendering the network robust to unclean data.
%that is robust to unclean data by weakening the influence of the back-propagated gradients caused by wrong labels.
The proposed TSMAN consists of two sub-networks that are connected by three types of attention models in different layers.
The target of each attention model is to indicate potentially incorrect gradients in a certain layer for both sub-networks by analyzing their inferred features using the same input.
In order to achieve this purpose, the attention models are designed based on the propagation analysis of noisy gradients at different layers. This allows the attention models to effectively discover incorrect labels and weaken their influence during the parameter updating process.
%layers to make them effective in discovering the wrong labels and weakening their influences on parameter updating. 
By exchanging multi-level features within the two-stream architecture, the effects of noisy labels in each sub-network are reduced by decreasing the updating gradients.
Furthermore, a hierarchical distillation is developed to provide more reliable pseudo labels for unlabelded data, which further boosts the performance of our retrained TSMAN.
The experiments using both the HVSMR 2016 and BRATS 2015 benchmarks demonstrate that our semi-supervised learning framework surpasses the state-of-the-art fully-supervised results. 
\end{abstract}

\section{Introduction}
\label{sec:intro}
Recently, many successful deep networks have been proposed to segment 3D magnetic resonance (MR) data \cite{Yu2017,Tseng_2017_CVPR,Cicek2016,Yu2017,Yu2017a}.
However, the scarcity of clean, labeled data severely hinders further development of deep learning methods for real applications.
Even for manual annotation, it is inevitable that even experts may make mistakes due to the effects of fatigue and human error.
%for experts to make some negligence due to exhaustion.
Thus, it is urgently important to improve the robustness of networks to noisy labels and generate more reliable machine annotations.
%improving the robustness of networks to noisy labels and generating more reliable annotations by machine are both important and urgent.
In this paper, we design a network that is less disturbed by noisy labels and propose a simple but effective distillation model to generate reliable pseudo labels.

Self-training is a typical semi-supervised method, which generates pseudo labels for unlabeled data by using trained models.
Obviously, the quality of pseudo labels is crucial to the performance of a final retrained model.
Among existing methods that generate labels automatically, model distillation~\cite{Hansen1990,Gupta2016} is one of the most widely used methods, which aggregates the inferences from multiple models for better pseudo labels.
Different from model distillation, Radosavovic \emph{et~al.} \shortcite{Radosavovic_2018_CVPR} recently proposed a data distillation that aggregates the inferences from multiple transformations of a data sample; this method proves to be superior to model distillation.
Although both distillation methods are effective, the generated pseudo labels are still noisy, which limits the performance of self-training.

\begin{figure}[t]
	\begin{center}
		\includegraphics[width=0.8\linewidth]{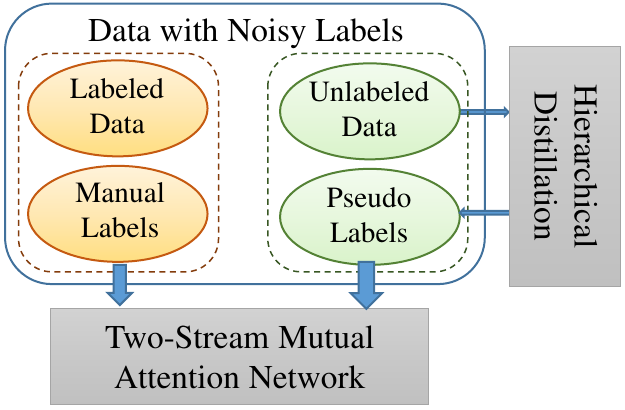}
	\end{center}
	\caption{The pipeline of our self-training framework. The hierarchical distillation first generates reliable pseudo labels for unlabeled data, and then mixed data is used to retrain the two-stream mutual attention network, which is robust to noisy labels. }
	\label{fig:pl}
\end{figure}

To resolve the problems caused by unclean data, Malach and Shalev-Shwartz~\shortcite{NIPS2017_6697} propose training two models and only updating them when their predictions are different.
They believe that the same predictions usually occur when two models obtain both right or wrong answers to easy or hard samples, respectively.
Thus, removing the hard samples with the same wrong predictions can effectively prevent incorrect updates, because annotations for hard samples are more likely to be noisy.
This inference procedure using prediction disagreement is useful for discovering noisy updates caused by incorrect labels; however, Malach and Shalev-Shwartz~\shortcite{NIPS2017_6697} ignore that the intermediate information during generating prediction is also important.
%In spite of promising performance on noisy classification dataset, Malach and Shalev-Shwartz~\shortcite{NIPS2017_6697} ignore the intermediate information of the inference procedure from two models that is also important to find out noisy updating caused by wrong labels.

In this paper, we propose a two-stream mutual attention network (TSMAN) by comprehensively exchanging multi-level features between two networks in different layers, including their predictions.
To give an example, our intuition tells us that if two students share a teacher, it is important to analyze the nature of these students mistakes to determine whether the errors are unique or result from instructional gaps.
%The intuition is that the analysis procedures of two students are important to find out the mistakes from their common teacher.
Based on this intuition, we use attention models in multiple layers to discover potential incorrect predictions and weaken the corresponding gradients during back-propagation. 
A vital challenge is how to provide useful clues about noisy gradients for attention models to infer noise distribution.
To address this issue, a two-stream architecture is developed by connecting two sub-networks with multi-attention models, which collect information from two sub-networks to discover noisy gradients.
By analyzing the noisy label propagation process, three kinds of attention models are designed for different layer depths, which successfully weakens the noisy gradients propagated by the loss layer.
By weakening the noisy gradients in multiple layers, our TSMAN is robust to noisy labels in biomedical data and performs comparably to fully-supervised learning methods when only partial annotations are used.

Furthermore, a hierarchical distillation method is proposed to combine data distillation \cite{Radosavovic_2018_CVPR} and model distillation \cite{Hansen1990}.
With the high quality of pseudo labels generated by our hierarchical distillation, the performance of TSMAN in self-training tasks can be further improved.

The whole self-training process of our method is shown in Figure~\ref{fig:pl}.
The overall contributions are summarized as follows:
\begin{itemize}
\item We propose a novel, two-stream mutual attention network that is robust to noisy labels and can be flexibly extended to many applications when clean annotations are difficult to acquire.

\item The proposed hierarchical distillation is more effective than either data or model distillation in generating reliable pseudo labels.

\item The proposed self-training segmentation framework with TSMAN and hierarchical distillation is superior to existing methods.

\end{itemize}

\section{Related Work}
\label{sec:related}

In this section, we briefly discuss two categories of related work: networks that are robust to noisy labels and self-training methods that use distillation.

In supervised learning models, the topic of improved resilience to noisy labels has been widely studied. 
Barandela and Gasca~\shortcite{Barandela2000} remove the labels that are suspected to be incorrect before retraining.
Inspired by the minimum entropy regularization in \cite{Grandvalet2005}, Reed et al.~\shortcite{Reed2014} propose adding a regularization term, which is related to the current prediction, to the network's loss function.
Mnih and Hinton~\shortcite{Mnih2012} use a probabilistic model to calculate the probability of each label being incorrect, and avoid updating in case incorrect labels occur.
McDowell~\emph{et~al.}~\shortcite{McDowell2007} propose novel generalizations for three comparison algorithms that examine how cautiously or aggressively each algorithm exploits noisy intermediate relational data.
Goldberger and Ben-Reuven~\shortcite{Goldberger2016} apply the Expectation Maximization (EM) algorithm by iteratively estimating true labels and retraining the network, which requires two-phase training to optimize two distinct softmax layers.
Recently, Malach and Shalev-Shwartz~\shortcite{NIPS2017_6697} tackle this problem by training two predictors with different initializations and only updating when there is a disagreement between their predictions.
Han~\emph{et~al.}~\shortcite{Han2018} present an effective co-teaching learning paradigm by simultaneously training two models and removing noisy samples from each mini-batch of data, which is conceptually similar to our method.
However, in our method for segmentation, each sample has densely arranged labels, which is different from the form of classification in ~\cite{Han2018}.
Thus, exploring the spatial relationship among labels helps us learn from noisy labels in the segmentation stage; this significant step is often ignored in existing methods.
Based on the recently success of attention mechanism \cite{he2018twofold,DBLP:conf/bmvc/YueMWZZ018,8237486,chen2018temporal}, we use the multiple attention models between layer pairs in two simultaneously trained networks to weaken noisy gradients. This process not only considers the prediction disagreement but also exchanges the evidence during inference.

The self-training approach is the earliest for semi-supervised learning, and it uses the predictions of a model on unlabeled data to retrain itself for better performance.
Without using post-processing, many distillation approaches have been widely adopted for self-training.
Gupta \emph{et~al.}~\shortcite{Gupta2016} propose the cross model distillation for tackling the problem of limited labels.
Laine and Aila~\shortcite{Laine2016} aggregate the inferences of multiple checkpoints during training to avoid training multiple models.
Besides model distillation, data distillation is also an effective method to explore new information from data transformations.
Simon \emph{et~al.}~\shortcite{Simon2017} obtain extra data from different views to retrain models, which yields an excellent performance in hand keypoint detection.
Moreover, Radosavovic \emph{et~al.}~\shortcite{Radosavovic_2018_CVPR} demonstrate that an inference to multiple transformations of a data point is superior to any of the predictions under a single transform.
Inspired by the above methods, we propose combining data distillations and model distillations in a hierarchical way to incorporate the unique advantages of both model and data distillation.
\section{Two-Stream Mutual Attention Network}
\label{sec:dan}
In this section, we first analyze the propagating process of noisy labels during training and then give a detailed description of the formulation and implementation of the two-stream mutual attention network.

\begin{figure*}[t]
	\begin{center}
		\includegraphics[width=6.8in]{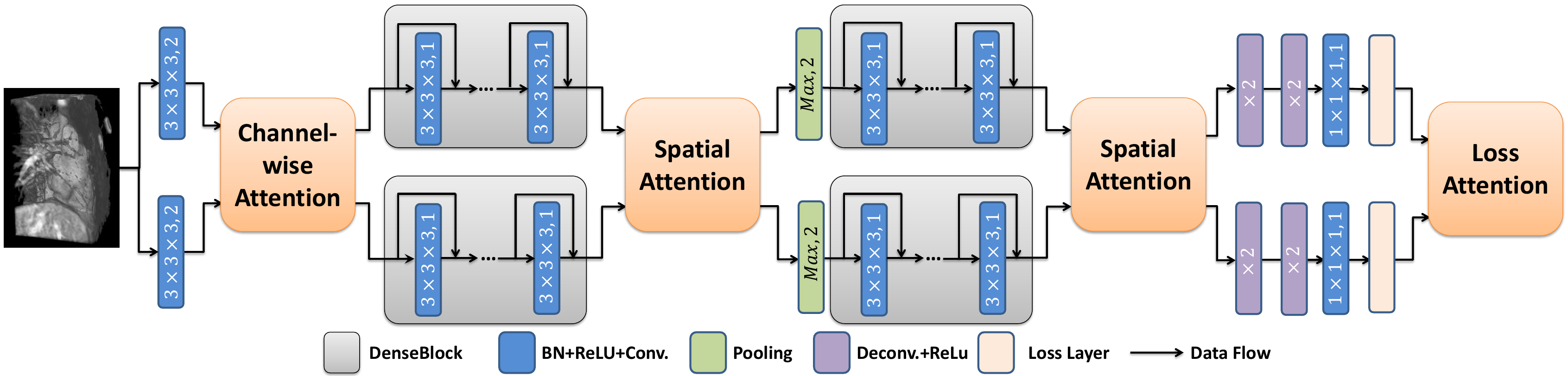}
	\end{center}
	\caption{The diagram of TSMAN. For readability, we present the parameters of Conv, pooling, and Deconv layers on the operation units. Each DenseBlock consists of $12$ BN+ReLU+Conv units.}
	\label{fig:dan}
\end{figure*}
\subsection{Problem Formulation}
We denote $\boldsymbol{x}$ as an input data sample with $N$ elements and $\tilde{\boldsymbol{y}}$ as the observed labels, possibly with some noise. For example, in a binary segmentation task, $\boldsymbol{x}$ can be a 2D array with $N$ pixels, and $\tilde{\boldsymbol{y}}=\{\pm 1\}^{N}$ is the label array indicating which class each pixel belongs to. Our goal is to train a model on $\{\boldsymbol{x},\tilde{\boldsymbol{y}}\}$ that is comparable to $\{\boldsymbol{x},\boldsymbol{y}\}$, where $\boldsymbol{y}$ is the clean label array. For simplicity, we denote $\boldsymbol{y}=\{\pm 1\}^N$. Then, $\tilde{\boldsymbol{y}}$ can be written as $\tilde{\boldsymbol{y}}=\bm{\theta}.*\boldsymbol{y}$, where $.*$ denotes element-wise multiplication of two arrays. For most pixels whose labels are correct, $\theta_i=1$. $\theta_i= -1$ indicates that the label is wrong for pixel $i$. 

Without losing generalization, a fully convolutional network is trained on $\{\boldsymbol{x},\widetilde{\boldsymbol{y}}\}$, in which the weights updating layer $d$ are represented by:
\begin{eqnarray}\label{eq:lw}
\frac{\partial L_s(\boldsymbol{p},\bm{\theta}.*\boldsymbol{y})}{\partial \boldsymbol{w}^d} &=& \frac{\partial L_s(\boldsymbol{p},\bm{\theta}.*\boldsymbol{y})}{\partial \boldsymbol{o}^{d}}\frac{\partial\boldsymbol{o}^{d}}{\partial\boldsymbol{w}^{d}},
\end{eqnarray}
\begin{eqnarray}\label{eq:lo}
\frac{\partial L_s(\boldsymbol{p},\bm{\theta}.*\boldsymbol{y})}{\partial \boldsymbol{o}^d} &=& \frac{\partial L_s(\boldsymbol{p},\bm{\theta}.*\boldsymbol{y})}{\partial \boldsymbol{o}^{d+1}}\frac{\partial\boldsymbol{o}^{d+1}}{\partial\boldsymbol{o}^{d}},
\end{eqnarray}
where  $\boldsymbol{o}^d$ is the output features of layer $d$, $\boldsymbol{w}^d$ is the convolutional weights, $L_s$ is the objective function, and $\boldsymbol{p}$ is the prediction of the network.
Therefore, the target is to weaken the gradients from $y_i$ with $\theta_i=-1$ in Eq.~\eqref{eq:lw} and Eq.~\eqref{eq:lo}.

To this end, an intuitive way is to obtain an attention model $f_{att}$, and then:
\begin{eqnarray}\label{eq:at}
\frac{\partial L_s(\boldsymbol{p},\bm{\theta}.*\boldsymbol{y})}{\partial \boldsymbol{p}} &\Rightarrow& \frac{\partial L_s(\boldsymbol{p},\bm{\theta}.*\boldsymbol{y})}{\partial \boldsymbol{p}}f_{att}(\boldsymbol{p},\boldsymbol{h}).
\end{eqnarray}
Ideally, when $\theta_i=-1$, $f_{att}(p_i,\boldsymbol{h})$ is expected to be $0$ with the extra information $\boldsymbol{h}$.
Next, we will introduce how to provide useful extra information $\boldsymbol{h}$ to indicate the potential noisy gradients in the network.

\subsection{Two-Stream Architecture}
We believe that the inference processing of another network is helpful to discover incorrect updates in this network.
Thus, we train two networks with the same inputs and use the predictions $\widehat{\boldsymbol{p}}$ from the other network as $\boldsymbol{h}$ in Eq.~\eqref{eq:at}:
\begin{eqnarray}\label{eq:at3}
\frac{\partial L_s(\boldsymbol{p},\bm{\theta}.*\boldsymbol{y})}{\partial \boldsymbol{o}^d} &\Rightarrow& \frac{\partial L_s(\boldsymbol{p},\bm{\theta}.*\boldsymbol{y})}{\partial \boldsymbol{o}^d}f_{att}^k(\boldsymbol{p},\widehat{\boldsymbol{p}})
\end{eqnarray}

However, as the information in $\widehat{\boldsymbol{p}}$ is unable to indicate all wrong gradients in $\frac{\partial L_s(\boldsymbol{p},\bm{\theta} \boldsymbol{y})}{\partial \boldsymbol{o}^d}$, some noisy gradients will be propagated to previous layers.
To this end, multiple $f_{att}^k(\boldsymbol{o}^d,\boldsymbol{h}^d)$ are applied in different layers to weaken noisy gradients in the whole network as much as possible by:
\begin{eqnarray}\label{eq:at2}
\frac{\partial L_s(\boldsymbol{p},\bm{\theta}.*\boldsymbol{y})}{\partial\boldsymbol{o}^d} &\Rightarrow& \frac{\partial L_s(\boldsymbol{p},\bm{\theta}.*\boldsymbol{y})}{\partial \boldsymbol{o}^d}f_{att}^k(\boldsymbol{o}^d,\widehat{\boldsymbol{o}}^d).
\end{eqnarray}
Our two-stream mutual attention network is thus designed as a symmetric two-stream architecture, as shown in Figure~\ref{fig:dan}. The four attention models take in both feature maps from two sub-networks and two feedback attention maps to indicate their potential wrong gradients.
Three types of attention models are used: loss attention (LA), spatial attention (SA), and channel attention (CA). These attention models will be introduced in detail in the following section.

Finally, the loss of the TSMAN is:
\begin{eqnarray}\label{eq:fl}
L = L_s(\boldsymbol{p},\bm{\theta}.*\boldsymbol{y})+L_s(\widehat{\boldsymbol{p}},\bm{\theta}.*\boldsymbol{y})
\end{eqnarray}
The final prediction for a test sample is obtained by averaging the softmax outputs of two networks.

\subsection{Loss Attention}
\label{sec:a}
\begin{figure}[t]
	\begin{center}
		\includegraphics[width=3.4in]{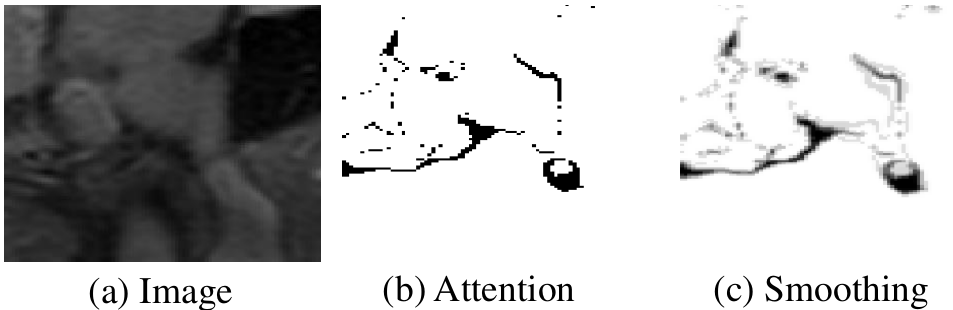}
	\end{center}
	\caption{The black pixels in (b) indicate that the predictions of two sub-networks for (a) are different. (c) is obtained by applying Gaussian smoothing to (b), which corresponds to Eq.~\eqref{eq:Q}. After smoothing, more voxels in the input image (a) are involved for back-propagation to improve the networks. }
	\label{fig:w}
\end{figure}

In our task, a reasonable hypothesis is that the same predictions from two sub-networks usually occur when the input sample is extremely simple or hard.
The extremely simple samples are easy to predict correctly by both sub-networks, which means that the loss can be ignored in back-propagation for model fine-tuning.
The extremely hard samples are more likely to be annotated falsely, which means that their labels are unreliable and can also be ignored in back-propagation.
Based on this hypothesis, the loss attention model (LA) is designed to remove these two kinds of samples by:
\begin{eqnarray}\label{eq:Q}
f_{att}^4(\boldsymbol{p},\widehat{\boldsymbol{p}}) =  \bm{\omega}(\boldsymbol{p} \oplus \widehat{\boldsymbol{p}})
\end{eqnarray}
where $\oplus$ is the voxel-wise exclusive OR operations, and $\bm{\omega}$ is the weights of a Gaussian smoothing convolution operation that is applied to $(\boldsymbol{p} \oplus \widehat{\boldsymbol{p}})$.
$f_{att}^4(\boldsymbol{p},\widehat{\boldsymbol{p}})$ serves as a binary loss selector.
Although checking disagreements between predictions effectively removes the voxels with the same $p_i$ and $\widehat{p}_i$, it ignores a special case when $p_i$, $\widehat{p}_i$ are both wrong but $\widetilde{y}_i$ is correct.
In this special case, the correct labels are useful but ignored in back-propagation by only using $\oplus$.
To this end, we introduce $\omega$ to alleviate this problem.
We observe that the disagreements of  $p_i$ and $\widehat{p}_i$ usually occur on the boundaries (black voxels in Figure~\ref{fig:w} (b)).
The voxels near these boundaries are challenging for sub-networks, but they are relatively easier for experts to annotate.
This means that an $x_i$ that is near boundaries is more likely to have both wrong $p_i$, $\widehat{p}_i$ and a correct $\widetilde{y}_i$.
Therefore, we employ a smoothing operation on the attention map to partially preserve the voxels near boundaries during back-propagation; the attention map for this is shown in Figure~\ref{fig:w} (c).
After smoothing, $f_{att}^4$ in Eq.~\eqref{eq:Q} becomes non-binary.

\label{sec:sca}

\subsection{Spatial and Channel-wise Attention}
Since loss attention only extracts information from the final predictions of two sub-networks, we also exploit the mutual information between the feature maps extracted by the two models. Two types of attention models are introduced: spatial attention models and channel-wise attention models.

\begin{figure}[t]
	\begin{center}
		\includegraphics[width=3.4in]{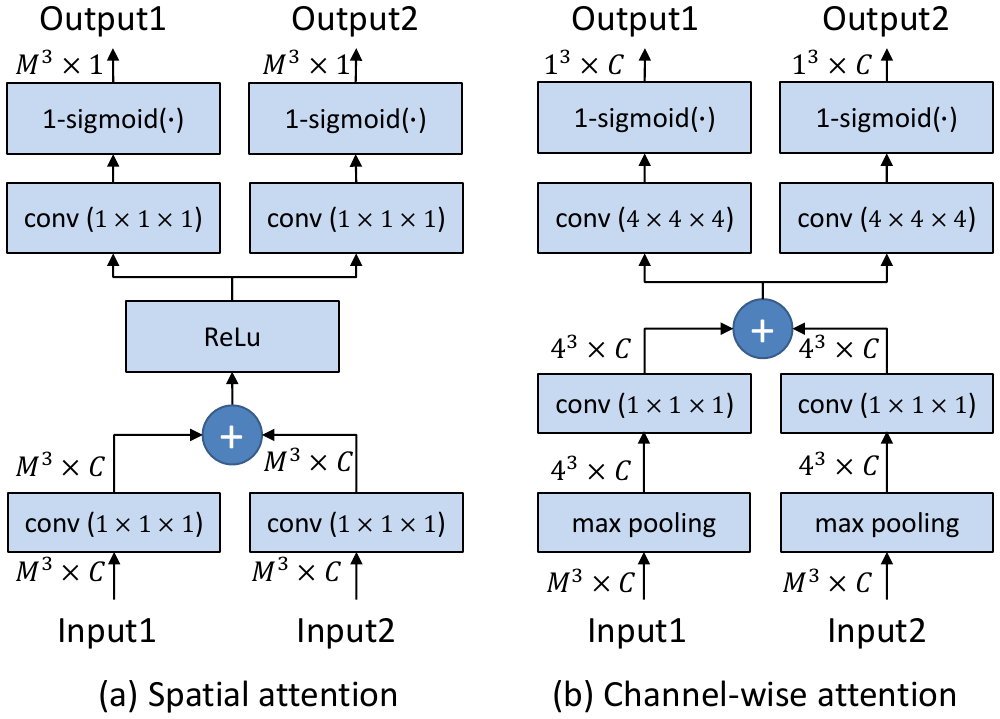}
	\end{center}
	\caption{The spatial attention (a) and channel-wise attention (b) diagrams. The blue circles indicate the element-wise additions, and the parameters in the convolution block indicate the kernel sizes. $M$, $C$ respectively represent the size and channel of the feature map. }
	\label{fig:csa}
\end{figure}

By defining $\boldsymbol{o}_{i}^d\in R^{C}$ as the feature vector at $i$-th position on the feature maps of layer $d$, where $C$ is the number of the feature map channel, we know  that the shallow layers receive more noisy gradients than deep layers, due to the larger receptive field on $\frac{\partial L_s(\boldsymbol{p},\bm{\theta} \boldsymbol{y})}{\partial \boldsymbol{o}^d}$. Furthermore, if $\boldsymbol{o}_{i}^d$ has a small d, it is more likely to receive a noisy gradient.
Based on these observations, we apply a spatial attention (SA) model to $f^{3}_{att}$ since it is close to the loss layer and few gradients of $\boldsymbol{o}_{i}^d$ are noisy.
Therefore, a spatial attention map is expected to weaken the noisy gradients in a small number of regions, which is efficient and feasible to implement.
For $f^{1,2}_{att}$, the attention models are too far from the loss layer, which means that the propagated gradient for almost all $\boldsymbol{o}_{i}^d$ have been polluted by Eq.~\eqref{eq:lo}.
Therefore, the spatial attention becomes inappropriate because the gradients of all feature vectors are noisy and should be weakened, which leads to a slow convergence.
To this end, the channel-wise attention (CA) is used for $f_{att}^{1,2}$ to select useful feature channels.
Both SA and CA serve as feature selectors during inference, as well as gradient selectors during back-propagation.
Figure~\ref{fig:csa} gives the detailed implementations of both SA and CA.

\label{sec:sca}
\section{Hierarchical Distillation}
\label{sec:hd}

While the proposed TSMAN provides an effective training strategy when there is noise in labels, a hierarchical distillation method is also proposed to reduce noisy labels ($\widetilde{y}_i$ with negative $\theta_i$) in pseudo labels. Our hierarchical distillation method integrates data distillation and model distillation together.
We define $\mathcal{L}$ as the labeled data space, $\mathcal{U}$ as the unlabeled data space, and
$f$ as the well-trained model on $\mathcal{U}$. The model distillation and data distillation respectively produce pseudo labels for $\mathcal{U}$ by:
\begin{eqnarray}\label{eq:md_dd}
P^{MD}(I) &=& g\big(\{f_t(I)|t=1,\cdots,T\}\big)\\
P^{DD}(I) &=& g\Big(\big\{\hbar^{-1}_{k}(f(\hbar_k(I)))|k=1,\cdots,K\big\}\Big)
\end{eqnarray}
where $I\in \mathcal{U}$ and $\hbar_{k}$ are the $k$-th transformation for $I$, which include rotation and flipping, while $\hbar_{k}^{-1}$ is the corresponding inverse transformation.
$g(\{\cdot\})$ is the voting function.
It is important to note that data distillation aggregates inferences of $K$ transformations of input data, which proves to be superior for model distillation.
However, this requires enough labeled data to train a suitable $f$.
Model distillation is more robust when the labeled data is insufficient, as it explores complementary information from $T$ models.

Both model and data distillation methods are effective in improving the reliability of pseudo labels for self-training, but they distill knowledge from different views.
Therefore, we combine them in a hierarchical way to take advantage of both of them:
\begin{eqnarray}\label{eq:hd}
P^{HD} &=& g\big(\{P_{t}^{DD}(I)|t=1,\cdots,T\}\big)
\end{eqnarray}
where $P_{t}^{DD}$ is the data distillation operation using $f_t$.
The experiments support our point that hierarchical distillation is superior to both data and model distillations individually.

\section{Experiments}
\label{sec:exp}
In this section, several ablation studies are first given to prove the effectiveness of the proposed method, and then comparisons with brand-new methods are introduced on HVSMR 2016 challenges \cite{Pace2015Interactive} and BRATS 2015 \cite{kistler2013virtual} benchmarks.

\subsection{Datasets}
The HVSMR 2016 dataset consists of $10$ 3D cardiac MR scans for training and $10$ scans for testing.
The resolution of each scan is about $200\times140\times120$.
All the MR data is scanned from patients with congenital heart disease (CHD), which is hard to diagnose.
The annotations contain the myocardium and blood pool regions in cardiac MR images.
The testing results are submitted to a public platform and evaluated by the organizer.
To alleviate the problem of overfitting, we apply data augmentations, including random rotations and flipping.

The training set for the BRATS-2015 dataset consists of 220 subjects with high-grade gliomas and 54 subjects with low-grade gliomas.
The resolution of each MRI image is $155\times 240\times 240$.
The platform for BRATS-2015 requires disguised evaluation, and most methods have fully-supervised training without published experimental settings. 
Thus we follow \cite{Tseng_2017_CVPR} by using 195 high-grade gliomas and 49 low-grade gliomas in the training set, and the remaining 30 subjects for evaluation.
There are five labels that correspond to common issues: edema, non-enhancing core, necrotic core, and enhanced core regions.

\subsection{Evaluation Metrics}
For HVSMR 2016, we use the overall score (higher is better) provided by the official platform for ablation analysis as our evaluation metric.
In comparison with recent methods from HVSMR 2016, we report three main metrics from the platform, including the mean Dices (a higher value is better), the average distance of boundaries, and the Hausdorff distance (lower values are better).
For BRATS-2015, we report the mean Dices criterion for all the five labels.

\subsection{Implementation Details}
We employ the DenseVoxNet \cite{Yu2017} as the sub-network in our two-stream architecture.
The training parameters, as well as the data pre-processing, follow the settings in \cite{Yu2017}, except for the max iterations, which are $35,000$ due to the disturbance of noisy labels.

For hierarchical distillation, $12$ geometric transformations are applied to each data point, including combinations of four rotations and three flips.
Three models, with different initializations and max iterations ($10000$, $15000$ and $20000$), are used for the ensemble.

In order to evaluate the robustness of experimental methods to noisy labels, the training datasets are divided into two parts; the labels for these parts are provided manually and by our hierarchical distillation.
The proportions of manual labels are controlled by $\xi$ to imitate different situations of noisy labels.

\begin{table}[h]
\footnotesize
\begin{center}
\caption{Comparison of the overall scores of different distillation methods on HVSMR 2016 with $\xi=30\%$, $50\%$, and $80\%$. $DD$ indicates the data distillation, and MD is the model distillation. A, B, and C are three base models to be aggregated.} \label{tab:hd-score}
\begin{tabular}{c|p{0.7cm}|p{0.7cm}|p{0.7cm}|p{0.7cm}}
  \hline
  $\xi$&A&B&C&MD \\
  \hline
  $30\%$&0.005&0.154&0.171&0.184\\
  $30\%$(DD)&0.104&0.163&0.168&$\mathbf{0.191}$\\
  \hline
  $50\%$&0.136&0.279&0.350&0.359\\
  $50\%$(DD)&0.196&0.33&0.356&$\mathbf{0.387}$\\
  \hline
  $80\%$&0.743&0.764&0.797&0.771\\
  $80\%$(DD)&0.785&0.775&0.806&$\mathbf{0.810}$\\
  \hline
\end{tabular}
\end{center}
\end{table}

\begin{table}[h]
\footnotesize
\begin{center}
\caption{Effects of different attention models. The baseline is our TSMAN without any $f_{att}$.} \label{tab:dan}
\begin{tabular}{c|p{0.7cm}|p{0.7cm}|p{0.7cm}|p{0.7cm}|p{0.85cm}}
  \hline
  &$f_{att}^1$&$f_{att}^2$&$f_{att}^3$&$f_{att}^4$&score \\
  \hline
  Baseline&&&&&-1.167\\
  \cline{1-6}
  \multirow{4}*{TSMAN}&&&&$\surd$&-0.737\\
  \cline{2-6}
  &&&$\surd$&$\surd$&-0.601\\
  \cline{2-6}
  &&$\surd$&$\surd$&$\surd$&-0.576\\
  \cline{2-6}
  &$\surd$&$\surd$&$\surd$&$\surd$&$\mathbf{-0.561}$\\
  \hline
\end{tabular}
\end{center}
\end{table}
\subsection{Evaluation of Hierarchical Distillation.}
In this part, we compare our hierarchical distillation with data distillation \cite{Radosavovic_2018_CVPR} and model distillation \cite{Hansen1990} under different settings.
The implementations of data and model distillation are special cases of hierarchical distillation that use a single model or no geometric transformation.
In Table~\ref{tab:hd-score}, we report the performance of the above methods when we split the HVSMR dataset with $\xi=30\%$, $50\%$, and $80\%$.

From Table \ref{tab:hd-score}, it can be observed that both data and model distillations are effective in improving the quality of pseudo labels.
However, when $\xi=30\%$, model distillation performs better than data distillation according to the overall scores $0.184$ and $0.168$ (highest among three base models with data distillation) respectively. The opposite case is true when $\xi=80\%$.
A possible reason for this is that the performance of multi-transformation inferences relies on the ability of its base models.
When the base models are trained with insufficient labels ($30\%$), the data distillation improvements are weak or even negative, as in the case of model C.
However, model distillation is less dependent on a certain model, as it distills knowledge from multiple models.
Therefore, it can be concluded that data distillation is more effective when base models are well-trained with plenty of correct labels, while model distillation is more robust to insufficient clean labels.
Both advantages of data and model distillation are crucial due to the various situations of biomedical datasets.
Therefore, our hierarchical distillation is a more general method as it takes advantage of both model and data distillations.

The pseudo labels generated by the hierarchical distillation are used to retrain our TSMAN, as well as other methods in comparison, in the following experiments.

\subsection{Analysis of the TSMAN}
In order to demonstrate the effectiveness of the TSMAN'S proposed design, we first explore the effects of different attention models in TSMAN.
As noisy gradients are propagated from deep to shallow layers, we report the results by 
sequentially applying attention models from $f_{att}^{4}$ to $f_{att}^{1}$ in Table~\ref{tab:dan}. 
From the information in Table~\ref{tab:dan}, adding a $f_{att}^{k}$ $(k=1,2,3,4)$ each time obviously improves the performance, which means that all $f_{att}^{1,2,3,4}$ effectively weaken noisy gradients in their corresponding layers. 
Besides, we find that the improvement of $f_{att}^k$ is usually weaker than $f_{att}^{k+1}$.
The reason is that noisy gradients in shallow layers are harder to discover, and some noisy gradients have been weakened by the latter attention model.

\begin{figure}[t]
	\begin{center}
		\includegraphics[width=3.4in]{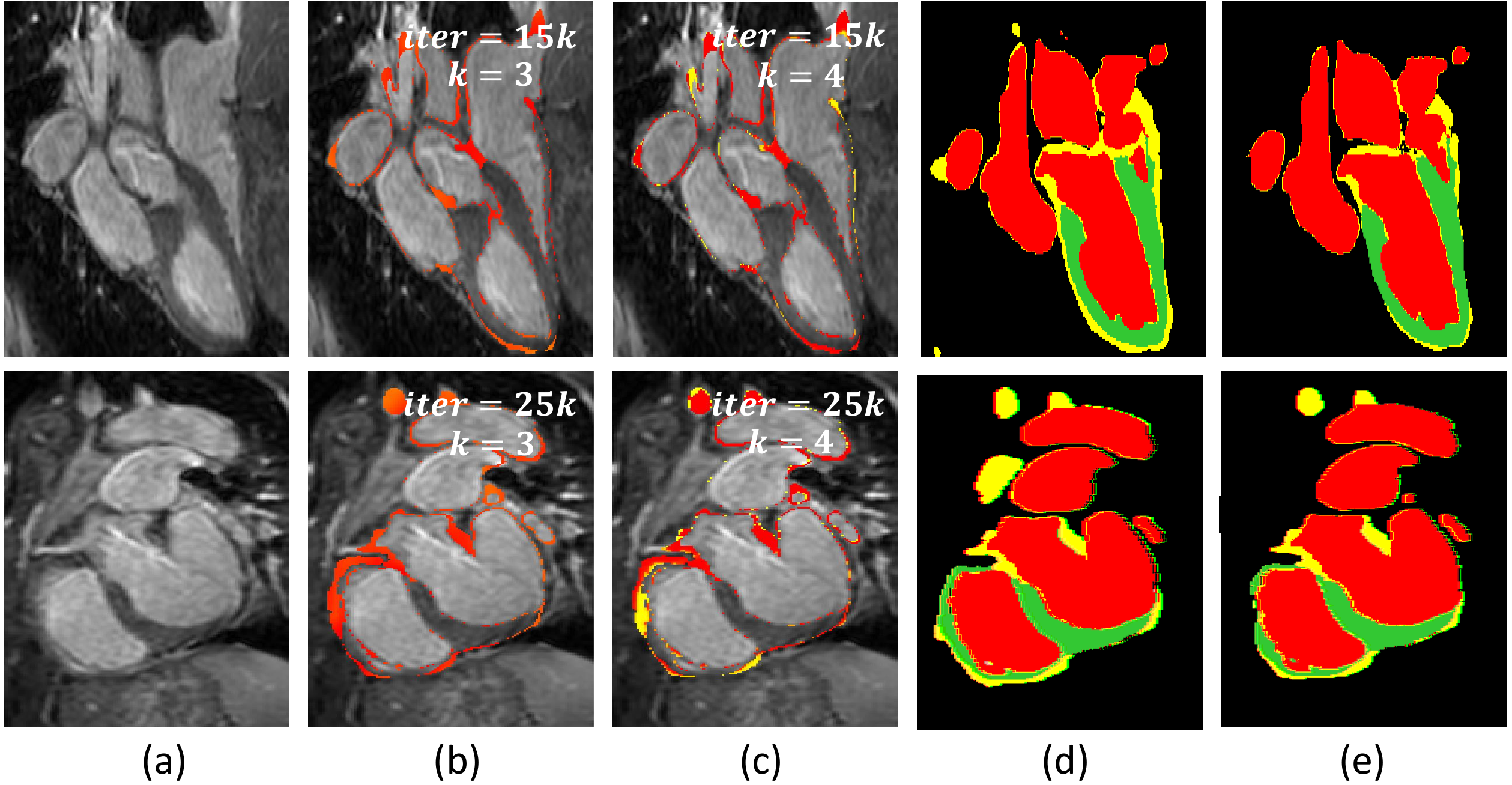}
	\end{center}
	\caption{Two sections of 3D MRI data. The labels in the red regions in (b), (c) are wrong, and the gradients in the yellow regions are weakened by $f_{att}^3$ and $f_{att}^4$, respectively. Notably, the more yellow, the smaller the response in the attention map. (d) and (e) are the predictions before and after retraining using unlabeled data. The red regions are the myocardium, the green regions are blood pools, and the yellow regions are incorrect predictions.}
	\label{fig:results}
\end{figure}
Next, some visual examples of spatial attentions are shown in Figure~\ref{fig:results} (b) and (c) to prove the effects of $f_{att}^{3}$ and $f_{att}^{4}$ in weakening noisy gradients.
From Figure~\ref{fig:results} (b) and (c), we observe that partial noisy gradients, marked in yellow, have been correctly eliminated by attention models, which proves the effectiveness of multiple attention models.
Especially for $f_{att}^{4}$, we also explore the effects of smoothing operations in Eq.~\ref{eq:Q}, and an 0.1 score improvement is obtained by using $\bm{\omega}$ with a kernel size of $3\times 3$ and a variance of $0.5$.
Finally, after retraining, the segmentation of TSMAN is obviously better, as shown in Figure~\ref{fig:results} (d) and (e).

\subsection{Comparisons Using the HVSMR 2016 Dataset}
\begin{figure}[t]
	\begin{center}
		\includegraphics[width=3.4in]{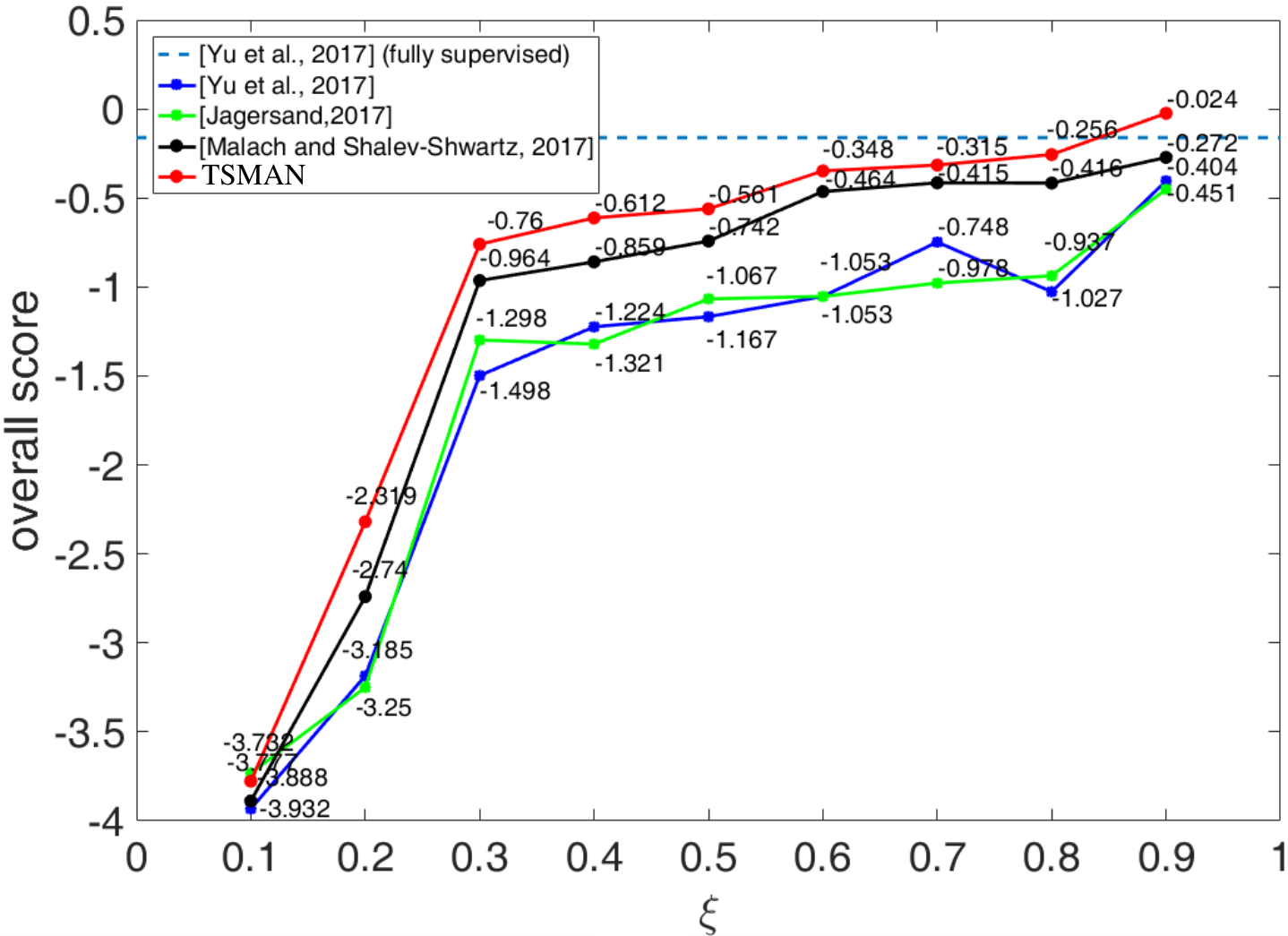}
	\end{center}
	\caption{Evaluations of different methods on HVSMR 2016 dataset with different $\xi$. The overall scores are calculated using the testing data.}
	\label{fig:curve}
\end{figure}

In order evaluate the robustness of our method to different unclean datasets, we compare the TSMAN with state-of-the-art methods \cite{NIPS2017_6697,Jagersand2017,Yu2017} by splitting the HVSMR 2016 dataset with $\xi=10\%, \cdots,90\%$.
\cite{NIPS2017_6697} is implemented by training two networks and assigning a non-zero loss weight for a voxel when their predictions are different.
For a fair comparison, we train two models for the other methods, and the predictions are averaged for the final results.
The results are shown in Figure~\ref{fig:curve}.
As illustrated, when the labeled data is insufficient ($\xi\in[10\%,20\%]$), the performances of all the learning-based methods are unsatisfactory.
This is because the quality of pseudo-labels is insufficient for providing useful information for model retraining.
When $\xi$ increases, noise decreases in pseudo-labels and models begin to learn extra knowledge from the unlabeled data.
Then, the performance of our TSMAN~and \cite{NIPS2017_6697} is better than \cite{Jagersand2017} and the baseline.
The reason is that student models learn more from fewer noisy labels, thus they have more consistent opinions during learning, which filters out many noisy labels.
Note that our TSMAN~is better than \cite{NIPS2017_6697}, as shown in Figure~\ref{fig:curve}, which demonstrates that the exchange of multi-level features is better than only the predictions.
Finally, when $\xi=0.9$, our retrained model even surpasses the fully-supervised model, which proves that there is also noise in the manual labels.

Next, we compare our TSMAN (using $\xi=0.9$) with the state-of-the-art methods on the leaderboard of the HVSMR 2016 benchmark, which are all trained in a fully supervised manner.
Table~\ref{tab:fr} shows the comparison results.
It should be noted that DenseVoxNet (Y17) is the baseline and our two-stream mutual attention network improves the ranking of DenseVoxNet from 3rd to 1st.
Compared with DenseVoxNet (Y17), we obtain a significant improvement on the ADB and HDD metrics, which demonstrates that TSMAN is more robust to noisy labels in the boundary regions.
Furthermore, the results demonstrate that the provided labels in the HVSMR 2016 dataset are not completely clean, so total trust in labels may be dangerous.

\begin{table}[h]
\footnotesize
\begin{center}
\caption{Comparison of different approaches using the HVSMR 2016 dataset. To save space, we use the first initial of the first author's last name combined with the last two digits of the year to indicate the methods, which respectively refer to [Mukhopadhyay, 2016], [Tziritas, 2016], [Van Der Geest, 2017], [Wolterink et al., 2016],  [Yu et al., 2016] and [Yu et al., 2017] from top to bottom.}
\label{tab:fr}
\begin{tabular}{p{0.5cm}|p{0.7cm}|p{0.7cm}|p{0.5cm}|p{0.7cm}|p{0.7cm}|p{0.5cm}|p{0.9cm}}
  \hline
  &\multicolumn{3}{c|}{Myocardium} &\multicolumn{3}{c|}{Blood Pool}&Overall\\
  \cline{2-7}
  &Dice&ADB&HDD&Dice&ADB&HDD&Scores\\
  \hline
  M16 &0.495&2.596&12.8&0.794&2.550&14.6&NA\\
  \hline
  T16 &0.612&2.041&13.2&0.867&2.157&19.7&-1.408\\
  \hline
  V17&0.747&1.099&5.09&0.885&1.553&9.41&-0.330\\
  \hline
  W16&0.802&0.957&6.13&0.926&0.885&7.07&-0.036\\
  \hline
  Y16&0.786&0.997&6.42&$\mathbf{0.931}$&$\mathbf{0.868}$&$\mathbf{7.01}$&-0.055\\
  \hline
  Y17&$\mathbf{0.821}$&0.960&7.29&$\mathbf{0.931}$&0.938&9.53&-0.161\\
  \hline
  Ours&0.820&$\mathbf{0.824}$&$\mathbf{4.73}$&0.926&0.957&8.81&$\mathbf{-0.024}$\\
  \hline
\end{tabular}
\end{center}
\end{table}

\subsection{Comparisons using the BRATS 2015 dataset}
\begin{figure}[t]
	\begin{center}
		\includegraphics[width=3.4in]{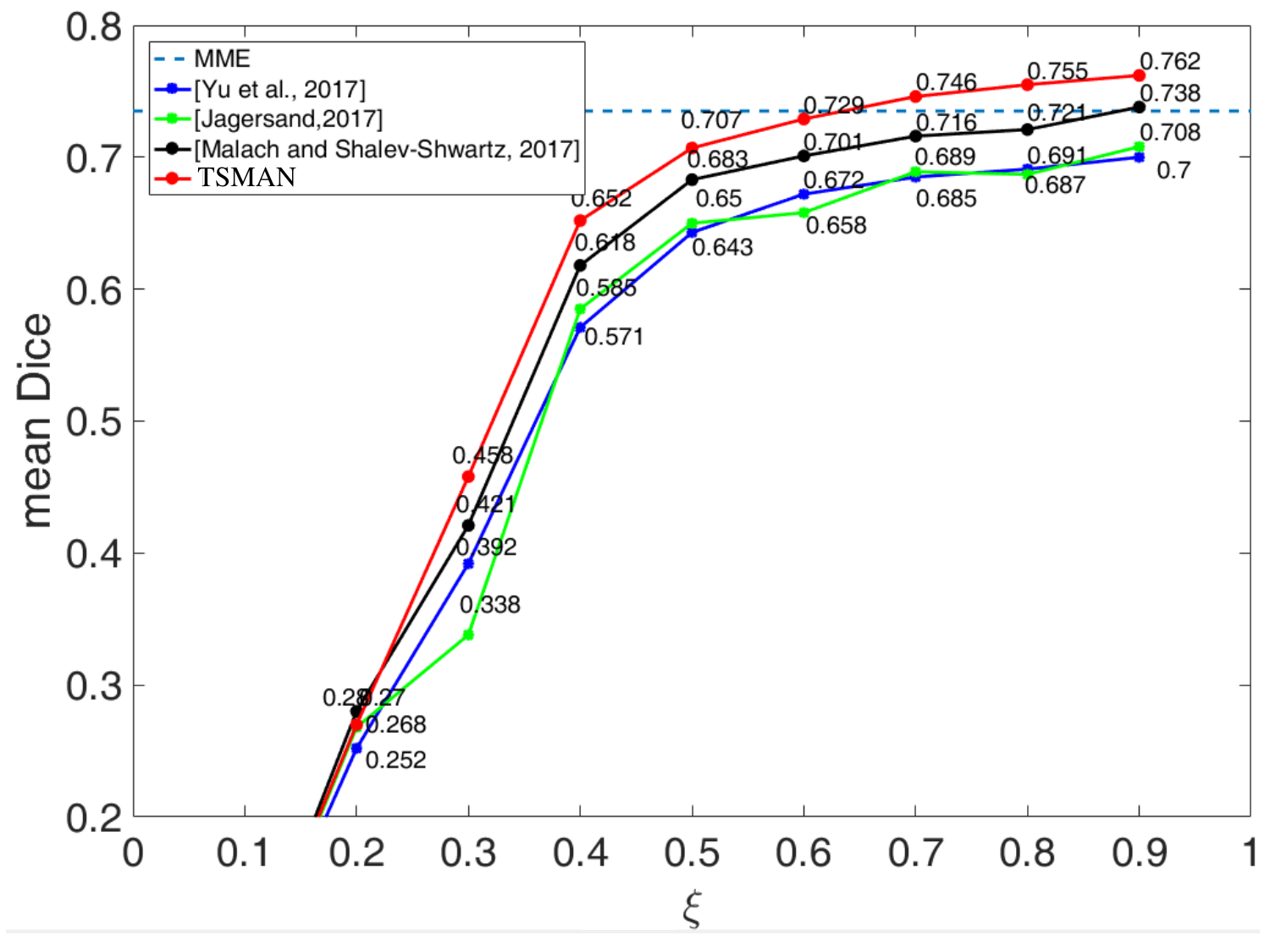}
	\end{center}
	\caption{Evaluations of different methods on BRATS 2015 with different $\xi$. The mean Dice of five labels are reported in the validation set.}
	\label{fig:curve1}
\end{figure}

\begin{table}[h]
\footnotesize
\begin{center}
\caption{Comparison of recent approaches on the BRATS 2015 dataset.}
\label{tab:brats}
\begin{tabular}{p{1cm}|p{0.7cm}|p{0.7cm}|p{0.7cm}|p{0.7cm}|p{0.7cm}|p{0.7cm}}
  \hline
  Label& 0&1&2&3&4&mean\\
  \hline
  \hline
  U-Net&0.923&0.429&$\mathbf{0.736}$&0.453&0.620&0.632\\

  MME&0.966&$\mathbf{0.943}$&0.712&0.328&$\mathbf{0.960}$&0.782\\
  DVN&0.989&0.426&0.730&0.645&0.850&0.728\\
  TSMAN&$\mathbf{0.990}$&0.7760&0.720&$\mathbf{0.684}$&0.790&$\mathbf{0.792}$\\
  \hline
\end{tabular}
\end{center}
\end{table}
We further evaluate our method on a larger 3D MRI dataset, the Brats 2015 benchmark, to prove its effectiveness.
Using experiments similar to those with the HVSMR 2016 dataset, we compare TSMAN with \cite{NIPS2017_6697}, \cite{Jagersand2017}, MME \cite{Tseng_2017_CVPR} and baseline DVN\cite{Yu2017} with different $\xi$.
Notably, we use the public results of MME and 3D U-net in \cite{Tseng_2017_CVPR} that use one-phase training, and the mean IOU for five labels are reported.

Figure~\ref{fig:curve1} gives evaluations of different methods using the BRATS 2015 dataset with different $\xi$.
From the results, it is apparent that the performance of TSMAN and \cite{NIPS2017_6697} are better than \cite{Jagersand2017} and baseline DenseVoxNet.
This further proves that improving robustness to noisy labels obviously benefits the semi-supervised learning performance.
Especially since Tseng  \emph{et~al.} \cite{Tseng_2017_CVPR} only provide the results of fully-supervised models without codes available, we use this as a fully supervised baseline.
Finally, when $\xi=0.9$, our retrained model also surpasses the fully-supervised \cite{Tseng_2017_CVPR}.

In Table~\ref{tab:brats}, we compare our TSMAN model ($\xi=0.9$) with U-Net \cite{Ronneberger2015}, MME \cite{Tseng_2017_CVPR}, and DVN \cite{Yu2017}, which are fully-supervised and trained.
This also demonstrates that TSMAN is robust to noisy labels.

\section{Conclusion}
In this paper, we propose a two-stream mutual attention network (TSMAN) that is robust to noisy labels. This network discovers incorrect labels and weakens the influence of these incorrect labels during the parameter updating process.
Specifically, three kinds of attention models are designed to connect multiple layers of two sub-networks; the attention models analyze the layers' features and indicate potential noisy gradients.
To improve the quality of pseudo labels, our hierarchical distillation takes advantage of both data and model distillation by hierarchically combining these two distillations.
Finally, combining TSMAN and hierarchical distillation in a self-training manner leads to state-of-the-art performance on the HVSMR 2016 and Brats 2015 benchmarks.

In the future, we hope that only one sub-network will be sufficient for completing inferences during testing. This may be achieved by generating attention maps and applying them to gradients of feature maps during the training process. 
Also, we will explore the effects of using different sub-networks, which may increase the challenge of designing attention models.

\section{Acknowledgement}
This work was supported by the National Natural Science Foundation of China under Nos. 61472377, 61632006, 91732304, and 61525206, the Fundamental Research Funds for the
Central Universities under Grant WK2380000002, the National Key Research and Development Program of China (2017YFC0820600), National Defense Science and Technology Fund for Distinguished Young Scholars (2017-JCJQ-ZQ-022).

\bibliographystyle{named}
\bibliography{dan}

\end{document}